# Scaling HuBERT for African Languages: From Base to Large and XL

Antoine Caubrière & Elodie Gauthier
Orange Research, France
`firstname.lastname@orange.com`

## 1 Introduction

Despite recent progress in multilingual speech processing such as in [1], [2] or [3], African languages remain under-represented in both research and deployed systems, particularly when it comes to strong, open-weight encoders that transfer well under low-resource supervision. Self-supervised learning has proven especially promising in such settings, yet most publicly released models targeting African speech remain at BASE scale, leaving unanswered whether larger encoders, trained exclusively on Africa-centric audio, offer tangible benefits and how model capacity interacts with data composition. This work addresses that gap by introducing `SSA-HuBERT-Large` (317M parameters) and `SSA-HuBERT-XL` (964M parameters), the first large models trained solely on African speech, alongside a BASE size counterpart. We release these models as open weights[1]. By conducting a carefully controlled experimental study focused exclusively on Sub-Saharan languages, covering automatic speech recognition (ASR) and language identification (LID) tasks, we demonstrate that larger architectures significantly improve performance by effectively leveraging large audio datasets.

## 2 Experiments

**HuBERT Pre-training**  The pre-training data are those collected and pre-processed in [4]. This dataset comprise 60k distributed across the following languages, ordered by number of speech hours: Swahili; Hausa; Kinyarwanda; French[2]; Bambara; Lingala; Sango; Tamasheq; Maninkakan; Songhai; Fula; Luba-Lulua; Kituba; Zarma; Wolof; Dyula; Mossi; Gulmancema.

Using the Fairseq toolkit we first train a HuBERT model [5] following a BASE configuration[3] (400k steps; 4xA100; 8 accumulation steps; Batch Size of 93,75 seconds). We then use the learned representation to train larger architectures, as in the original paper, namely *Large* (450k steps; 4xH100; 32 acc steps; 56.25s BS) and *XL* (450k steps; 8xH100; 32 acc steps; 56.25s BS).

**Fine-tuning setup**  Supervised FT and evaluations were both make use of $FLEURS_{SSA}$ subset of FLEURS [6] (320h of read speech recordings in 20 Sub-Saharan African languages (mainly spoken in Eastern countries).

For ASR task, we fine-tuned our pretrained model using the CTC approach provided in the SpeechBrain toolkit. The model is augmented with two fully connected layers of size 1024, followed by a linear layer to match the output vocabulary. The fine-tuning process involves two stages: a joint fine-tuning on the whole $FLEURS_{SSA}$ subset (40 epochs; 1xA100; 4 samples per BS) and a transfer learning to create monolingual models (same settings). We evaluated the performance of our models against AfriHubert-n [7], a BASE-size model also adopting a HuBERT architecture, and XLS-R 128 [1], a LARGE-size model based on a wav2vec2 architecture. Each of these models were fine-tuned using the same configuration method as was employed for the creation of our own models.

To assess representation quality, we performed a LID task in two settings: (i) pooling on speech encoder output with softmax, and (ii) pooling with linear layers reducing features before softmax. Both ran for 15 epochs on 1xA100 and using a BS of 4. The LID performance of our models is compared solely to that of AfriHubert-n [7], as their approach is the most comparable to ours in terms of African languages coverage and architecture type, in contrast to XLS-R 128.

**Results**  Extensive evaluations were performed over the 20 languages of the $FLEURS_{SSA}$ subset by using a greedy decoding. It was observed that larger models consistently outperform their smaller counterparts and previous SSL models tailored for African languages. The findings emphasise the significance of model capacity in managing substantial and varied datasets (i.e. recording conditions and speech variability), with the XL model attaining the lowest character and word error rates across a wide range of languages in the transcription task. Scores are reported in Table 1.

---

[1] https://huggingface.co/collections/Orange/african-speech-foundation-models
[2] Only African-accented speech
[3] https://github.com/facebookresearch/fairseq/tree/main/examples/hubert/config/pretrain/

Table 1: CER and WER (%) obtained on the SSA test subset of FLEURS-102 dataset.

| | #Params | afr CER / WER | amh CER / WER | ful CER / WER | hau CER / WER | ibo CER / WER | kam CER / WER | lin CER / WER |
|---|---|---|---|---|---|---|---|---|
| SSA-HuBERT-base-v2 | 95M | 19.8 / 59.1 | 13.3 / 44.3 | 16.8 / 54.2 | 8.5 / 28.1 | 15.8 / 49.7 | 14.5 / 50.2 | 6.9 / 20.4 |
| AfriHuBERT-n | 95M | 12.0 / 36.9 | 10.8 / 35.5 | 17.2 / 55.8 | 9.0 / 30.5 | 13.8 / 45.1 | 13.7 / 48.4 | 6.4 / 19.1 |
| SSA-HuBERT-Large | 317M | 13.0 / 42.3 | **9.9 / 32.9** | **15.4 / 50.9** | 6.6 / 21.6 | 13.2 / 44.2 | 11.4 / 41.8 | 4.9 / 14.9 |
| XLS-R | 317M | **9.1 / 27.9** | **9.9 / 32.8** | 16.6 / 54.2 | 8.2 / 30.0 | 13.2 / 44.4 | 12.0 / 43.2 | 5.3 / 16.9 |
| SSA-HuBERT-XL | 964M | 12.4 / 39.8 | 10.3 / 34.3 | 16.7 / 52.7 | **5.5 / 19.6** | **12.8 / 43.3** | **10.7 / 39.7** | **4.3 / 13.6** |

| | | lug CER / WER | luo CER / WER | nso CER / WER | nya CER / WER | orm CER / WER | sna CER / WER | som CER / WER |
|---|---|---|---|---|---|---|---|---|
| SSA-HuBERT-base-v2 | 95M | 10.3 / 49.4 | 7.6 / 33.6 | 10.7 / 35.9 | 10.6 / 44.5 | 19.4 / 73.1 | 7.3 / 34.6 | 19.1 / 58.6 |
| AfriHuBERT-n | 95M | 9.7 / 48.0 | 6.9 / 30.1 | 8.8 / 29.8 | 9.1 / 40.4 | 19.0 / 68.7 | 6.0 / 29.5 | 16.3 / 51.2 |
| SSA-HuBERT-Large | 317M | 9.4 / 46.9 | 6.1 / 28.0 | 8.4 / **28.8** | 8.0 / 35.3 | **18.2 / 66.9** | 5.1 / 24.6 | 15.5 / 49.8 |
| XLS-R | 317M | 10.0 / 49.1 | 6.1 / 28.1 | 8.5 / 28.9 | 8.7 / 38.8 | 18.8 / 70.2 | 5.3 / 27.0 | 15.9 / 50.2 |
| SSA-HuBERT-XL | 964M | **9.0 / 45.6** | **5.8 / 27.0** | **8.0 / 33.7** | **7.0 / 32.7** | 18.3 / 67.7 | **4.7 / 23.2** | **15.3 / 49.2** |

| | | swa CER / WER | umb CER / WER | wol CER / WER | xho CER / WER | yor CER / WER | zul CER / WER | AVG CER / WER |
|---|---|---|---|---|---|---|---|---|
| SSA-HuBERT-base-v2 | 95M | 4.8 / 17.6 | 18.3 / 53.7 | 16.3 / 48.7 | 8.9 / 42.2 | 21.6 / 62.2 | 9.1 / 42.1 | *13.0 / 45.1* |
| AfriHuBERT-n | 95M | 5.3 / 20.9 | 17.0 / 50.5 | 15.0 / 45.4 | 7.4 / 36.6 | **18.6 / 54.4** | 7.2 / 36.0 | *11.5 / 40.6* |
| SSA-HuBERT-Large | 317M | 3.3 / 12.0 | 15.1 / **47.7** | 13.7 / 42.2 | 6.7 / 34.6 | 19.9 / 57.9 | 6.7 / 33.3 | *10.5 / 37.8* |
| XLS-R | 317M | 4.9 / 19.4 | 15.7 / 49.9 | 14.0 / 44.1 | 6.9 / 35.9 | 19.4 / 57.3 | 7.3 / 36.1 | *10.8 / 39.2* |
| SSA-HuBERT-XL | 964M | **2.7 / 10.1** | **14.6 / 50.6** | **12.4 / 40.0** | **6.3 / 33.5** | 19.0 / 55.9 | **6.2 / 31.0** | *10.1 / 37.2* |

<span style="color:purple">Purple</span> corresponds to seen languages (5) during pre-training iterations of our models (named **SSA-HuBERT-xx**)
<span style="color:blue">Blue</span> corresponds to unseen languages (15) during pre-training iterations of our models (named **SSA-HuBERT-xx**)

Not surprisingly, the performance gains are more pronounced for languages that have been seen during pretraining, highlighting the importance of data coverage during this step, but also the influence of the data volume in model effectiveness.

Table 2: LID Accuracy on the test set of FLEURS$_{SSA}$

| | AfriHuBERT-n | SSA-HuBERT-base-V2 | SSA-HuBERT-Large | SSA-HuBERT-XL |
|---|---|---|---|---|
| **LID** | 93.3 | 78.3 | 89.9 | 92.7 |
| **LID$_{smooth}$** | 93.5 | 85.5 | **93.7** | 93.1 |

As reported on Table 2, results are more mixed on the LID task. AfriHuBERT outperforms while utilising a more compact architecture. This could be explained by the diversity of languages used during pre-training, which broadens the model's generalisation capabilities, a key skill for this type of task.

## 3 Conclusion

The LARGE and X-LARGE models establish a new reference for scalable, Africa-centered SSL in speech, enabling improved ASR for under-represented languages and providing a foundation for downstream tasks such as spoken language understanding (SLU). The study highlights the critical role of data coverage. Expanding pre-training data to cover more languages and dialects, but also improving representation for morphologically rich languages (e.g. Amharic, Yorùbá) will likely yield further gains. Future work includes extending evaluation to complex downstream tasks (e.g., speaker diarization, SLU) and leveraging related open datasets (e.g., FLEURS-SLU) to push beyond ASR toward end-to-end speech understanding in African languages.